\documentclass[10pt,conference]{IEEEtran}
\IEEEoverridecommandlockouts
\usepackage{cite}
\usepackage{amsmath,amssymb,amsfonts}
\usepackage{algorithmic}
\usepackage{graphicx}
\usepackage{textcomp}
\usepackage{xcolor}
\usepackage{tcolorbox}
\usepackage{booktabs}
\usepackage{subcaption}
\def\BibTeX{{\rm B\kern-.05em{\sc i\kern-.025em b}\kern-.08em
    T\kern-.1667em\lower.7ex\hbox{E}\kern-.125emX}}
\begin{document}

\title{Automatic Generation of Multi-Objective Optimization Benchmark Problems via Evolutionary Algorithms and Large Language Models\thanks{This work was supported by the Joint Usage/Research Center for Interdisciplinary Large-scale Information Infrastructures (JHPCN) in Japan (Project ID: jh250063) and JSPS Grant-in-Aid for Scientific Research (B) (Grant Number: 24K03011).}}

\author{\IEEEauthorblockN{
Koki Onomura\IEEEauthorrefmark{1}, 
Yuto Araki\IEEEauthorrefmark{1},
Yuhiro Ono\IEEEauthorrefmark{2},
Tomohiro Harada\IEEEauthorrefmark{1}}
\IEEEauthorblockA{\IEEEauthorrefmark{1}Graduate School of Science and Engineering, Saitama University, Saitama, Japan \\
Email: \{onomura.k.200@ms., araki.y.017@ms., tharada@mail.\}saitama-u.ac.jp}
\IEEEauthorblockA{\IEEEauthorrefmark{2}Graduate School of Systems Design, Tokyo Metropolitan University, Tokyo, Japan\\
Email: ono-yuhiro@ed.tmu.ac.jp}}

\maketitle

\begin{abstract}
This study proposes a method that integrates large language models (LLMs) with evolutionary algorithms to automatically generate multi-objective optimization benchmark problems.
Existing mathematically defined benchmark problems often lack the diversity and specific characteristics found in real-world problems.
In contrast, the proposed method enables the generation of problems with diverse characteristics by leveraging the LLMs as genetic operators, such as crossover and mutation, to directly evolve problems represented as mathematical formulas.
To evaluate the effectiveness of the proposed method, we conducted experiments to generate benchmark problems in which either NSGA-II or MOEA/D performs better.
Experimental results showed that the proposed method generates benchmark problems possessing characteristics specific to each algorithm.
Specifically, NSGA-II-preferred problems exhibit significantly different scales between objective functions, which MOEA/D struggles with.
On the other hand, MOEA/D-preferred problems have convex Pareto fronts, for which decomposition-based search performs well.
These results demonstrate that the proposed method can autonomously discover benchmark problems that exploit the specific characteristics and weaknesses of each algorithm.
\end{abstract}

\begin{IEEEkeywords}
Multi-objective optimization, Evolutionary algorithms, Large language models, Benchmarking, Automated problem generation
\end{IEEEkeywords}

\section{Introduction}

Real-world optimization tasks often involve multiple conflicting objectives, such as simultaneously minimizing production costs while maximizing product quality in manufacturing.
These tasks are formulated as multi-objective optimization problems (MOPs)~\cite{Deb_2001}, in which the goal is not to obtain a single optimum but rather a set of non-dominated solutions, known as the Pareto optimal set (POS), that represent trade-offs among competing objectives. 

To solve MOPs, a wide variety of multi-objective evolutionary algorithms (MOEAs)~\cite{Coello_2007} have been proposed.
Representative algorithms include NSGA-II~\cite{DebAPM02} and MOEA/D~\cite{MOEAD_Zhang_Li}, which have been widely adopted in both academic and practical applications.
However, as stated by the No Free Lunch (NFL) theorem~\cite{Wolpert_Macready_1997}, no single algorithm can outperform all others on every possible problem.
Consequently, benchmark problem suites with diverse characteristics are essential for fair algorithm comparison and for understanding the strengths and limitations of different MOEAs~\cite{KUMAR2021100961}. 

Classic benchmark suites, such as ZDT~\cite{Deb2000} and DTLZ~\cite{1007032}, have played a central role in MOEA research.
Nevertheless, it has been argued that relying solely on specified test problems risks encouraging over-specialization to a narrow set of problem characteristics~\cite{PIOTROWSKI2023101378}.
As a result, algorithm performance on such benchmarks may not necessarily reflect robustness or generalizability to real-world problems with unknown structures~\cite{Zapotecas_Martinez_2019, Tanabe_Oyama_2017}.
Although benchmark problems derived from real applications could mitigate this issue, they are typically expensive to obtain, limited in number, and often cannot be publicly released due to confidential constraints.

From this background, there is a growing demand for methods that can efficiently generate diverse benchmark problems capable of evaluating algorithm performance from multiple perspectives.
In response to this need, recent work has proposed the Large Language Model-driven Evolutionary Benchmark Generator (LLM-EBG)~\cite{10.1145/3712255.3726663}, which employs a large language model (LLM) as a genetic operator within an evolutionary framework.

However, the existing LLM-EBG framework has been limited to single-objective optimization problems (SOPs) and has not yet been extended to MOPs.
In this study, we propose a multi-objective extension of LLM-EBG and investigate its effectiveness through a case study.
Specifically, our case study automatically generates MOP benchmark problems in which performance differences between 
NSGA-II and MOEA/D become especially pronounced.
This extension enables the automatic generation of diverse benchmark problems that reflect the strengths and weaknesses of different algorithms.
The proposed method contributes to more systematic algorithm characterization and supports the development of more robust MOEAs.

\section{Conventional Method: LLM-EBG} 

This study extends LLM-EBG~\cite{10.1145/3712255.3726663} to the MOP domain.
This section provides an overview of the LLM-EBG framework, which serves as the foundation of this study.

LLM-EBG is an automated optimization benchmark generation framework that integrates the advanced pattern processing capabilities of LLMs with the search capabilities of evolutionary algorithms (EAs).
In this framework, benchmark problems represented by mathematical expressions are regarded as individuals in an EA, and an LLM is used to perform variation operators such as crossover and mutation directly on these formula strings~\cite{Meyerson2024}.
By leveraging the natural language and pattern processing capabilities of LLMs as genetic operators for EAs, LLM-EBG enables direct crossover and mutation at the string level.

\begin{figure}[tb]
\begin{center}
\includegraphics[width=1\linewidth]{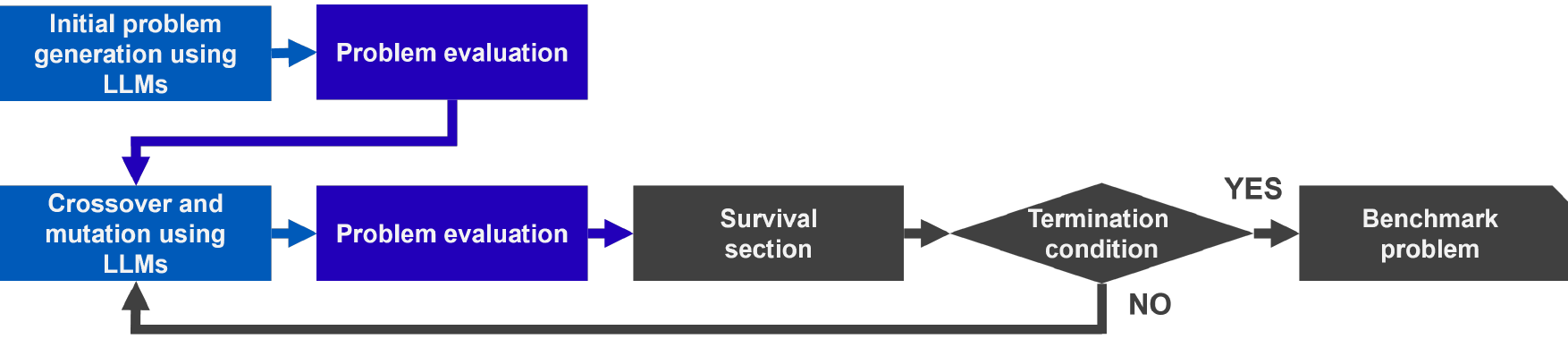}
\caption{Flow chart of LLM-EBG}
\label{fig:llm_ebg}
\end{center}
\end{figure}

Fig.~\ref{fig:llm_ebg} illustrates the flow of LLM-EBG, which follows a standard EA loop.
First, an LLM generates an initial population of benchmark problems, and each benchmark is evaluated using a predefined fitness function.
Subsequently, during the generational loop, offspring benchmarks are generated by the LLM through crossover and mutation and then evaluated.
Finally, the next generation is selected from the combined parent and offspring populations.

In LLM-EBG, genetic operators driven by an LLM constitute the core components of the framework.
Previous studies utilized few-shot prompting techniques \cite{chan2022datadistributionalpropertiesdrive}, in which example benchmark problems are included in the prompt together with explicit instructions for crossover or mutation, enabling the LLM to generate new benchmark problems.

Previous works focused on generating benchmark problems for single-objective optimization, either by emphasizing performance differences between algorithms \cite{10.1145/3712255.3726663} or by generating benchmarks that are difficult to optimize \cite{Araki2025SSI}.

\section{Proposed Method}
Although the LLM-EBG framework has successfully generated benchmark problems with desired characteristics, its application has so far been limited to single-objective optimization.
To overcome this limitation, this study extends LLM-EBG to automatically generate MOP benchmarks.

This section describes the components of the proposed framework.
Section~\ref{sec:method_expression} discusses the representation of benchmark problems.
Section~\ref{sec:method_llm} explains the crossover and mutation processes using an LLM, together with the prompt design.
Section~\ref{sec:method_fitness} explains the fitness design based on performance differences between algorithms.

\begin{figure}[tb]
  \centering
  \begin{tcolorbox}[
    colback=white,
    colframe=black!60,
    arc=2mm,
    boxrule=0.8pt,
    boxsep=3pt,
    left=4pt, right=4pt, top=4pt, bottom=4pt,
    fontupper=\scriptsize\ttfamily
  ]
You are an expert assistant for creating 
multi-objective optimization benchmark 
problems.

\vspace{1em}
\#\#\# Instruction \#\#\#

Generate a new multi-objective
optimization problem with exactly \{m\_obj\}
objective functions in \{D\} dimensions.
The objectives should be conflicting.

\vspace{1em}
You must follow these rules:

1. Provide the answer in terms of variables x[i], where i is from 0 to 4.
   
2. You can only use the operators: "+", "-", "*", "/", "**", "sqrt", "sin", "abs".
   
3. You must format the output strictly as follows, with each function on a new line:

   f1(x) = ...
   
   f2(x) = ...
   
4. Do not provide any other text, explanations, or constraints.

\vspace{1em}
\{indicator\}

\{formulas\}
  \end{tcolorbox}
  \caption{Input prompt format for the LLM}
  \label{fig:prompt_example}
\end{figure}

\subsection{Representation of Benchmark Problems} \label{sec:method_expression}

In the proposed method, a benchmark problem consisting of $M_{obj}$ objective functions is represented as a set of formula strings and regarded as a single individual.
For example, when $M_{obj}=2$, an individual $b$ is represented as a list of two formula strings $[f_1(x), f_2(x)]$.
This study targets minimization benchmark problems, where all objective functions are to be minimized.
Each objective function is represented as a string that can be evaluated using Python's \texttt{eval()} function.
Design variables are described as \texttt{x[0]}, \texttt{x[1]}, \dots, \texttt{x[D-1]} , forming a $D$-dimensional vector $\boldsymbol{x} = (x_0, \dots, x_{D-1})$.
Within the formula strings, arithmetic operations (\texttt{+}, \texttt{-}, \texttt{*}, \texttt{/}), exponentiation (\texttt{**}), and basic mathematical functions (\texttt{sqrt}, \texttt{sin}, \texttt{abs}) are defined as usable operators.

\subsection{Genetic Operations Using an LLM} \label{sec:method_llm}

Fig.~\ref{fig:prompt_example} shows the format of the prompt used during benchmark generation.
In the prompt, specific instructions are given together with multiple parent benchmark problems, role settings, situational descriptions, and constraint specification.
Here, \texttt{m\_obj} specifies the number of objectives, and \texttt{D} specifies the number of dimensions.
The fields \texttt{indicator} and \texttt{formulas} are given different instructions depending on the generation stages, as described below.
\begin{itemize}
\item \textbf{Initial population generation}: To ensure the diversity of the initial population, previously generated benchmarks are presented in \texttt{formulas}.
The \texttt{indicator} instructs the LLM as follows:  ``It is crucial to generate a diverse set of initial problems. The new problem should have different characteristics from the problems listed below. Consider using different mathematical structures, variable interactions, or function combinations.'' 
\item \textbf{Crossover}: The formulas of two parent benchmarks are presented in \texttt{formulas}, and the \texttt{indicator} instructs: ``Combine and create a new problem based on the following problems.''
\item \textbf{Mutation}: The formula of one parent benchmark is presented in \texttt{formulas}, and the \texttt{indicator} instructs: ``Slightly mutate and create a new problem based on the following problems.''
\end{itemize}
Through these operations, the genetic operators in LLM-EBG are realized.

During LLM benchmark generation, benchmarks that do not meet the required format or produce invalid evaluation results (e.g., division by zero, infinite values, or complex numbers) may be generated.
To eliminate such benchmarks, a pre-verification process is introduced. 
When a new benchmark is generated, $N_{val}$ (set to 10000 in this study) random inputs are sampled and evaluated using the objective functions in the benchmark.
If any evaluation value is invalid, the benchmark is considered unstable and rejected.
In addition, benchmarks with syntactic errors or with functions or variable names not permitted in the prompt---rendering them non-executable as Python code---are also rejected.
Rejected benchmarks are excluded from the population, and new benchmarks are regenerated.

\subsection{Evaluation of Generated Benchmark Problems} \label{sec:method_fitness}

A primary role of benchmark problems is to reveal the performance difference between optimization algorithms.
Accordingly, this study aims to automatically generate benchmarks in which a specified MOEA (hereinafter Algorithm A) outperforms a comparison MOEA (hereinafter Algorithm B). 
This performance difference is quantified and used as the fitness for evaluating generated benchmarks.

The fitness calculation proceeds as follows.
\begin{enumerate}
\item \textbf{Execution}: The benchmark $b$ is optimized independently $N_{trials}$ times each by Algorithms A and B.
\item \textbf{Normalization}: All non-dominated solution sets obtained from all $2 \times N_{trials}$ runs are collected to determine the ideal point $\boldsymbol{z}^*$ (minimum value for each objective) and the nadir point $\boldsymbol{z}^{nad}$ (maximum value for each objective) in the objective space.
All non-dominated solution sets are normalized to the range $[0, 1]$ using these points.
\item \textbf{HV Calculation}: The quality of each normalized Pareto front is evaluated using hypervolume (HV) \cite{Zitzler_Thiele_1999}.
The reference point is set to $\boldsymbol{r} = (1.1, 1.1)$, which lies outside the normalized objective space.
This yields two sets of HVs, $H_A$ and $H_B$, each containing $N_{trials}$ elements.
\item \textbf{Fitness Calculation}: Fitness is calculated based on the difference in rank sums, which serves as the basis for the Mann-Whitney U test.
The HV values of both algorithms are combined into a single set $H_{all} = H_A \cup H_B$, and ranks $rank(h) \in \{1, 2, \dots, 2N_{trials}\}$ are assigned in descending order of HV.
Specifically, the trial with the $n$-th largest HV is assigned $rank(h) = n$.
The fitness is defined as
\begin{equation}
fitness = \sum_{h \in H_B} \text{rank}(h) - \sum_{h \in H_A} \text{rank}(h).
\label{eq:fitness}
\end{equation}
\end{enumerate}

Fitness maximization encourages benchmarks in which Algorithm A consistently achieves higher HV values than Algorithm B, thereby providing a robust evaluation of performance superiority based on the entire HV distribution.

Although the pre-verification step eliminates invalid benchmarks, a benchmark without a trade-off relationship between objectives is not suitable for MOPs, even if it is evaluable.
Therefore, an additional post-verification step is introduced to confirm the existence of trade-offs.
During fitness evaluation, if the median HV of $H_{all}$ approaches the hypervolume of the reference point $\boldsymbol{r} = (1.1, 1.1)$ (i.e., 1.21)---specifically, if it exceeds 1.209---the Pareto front is regarded as having collapsed to a single solution. Since this indicates that no meaningful trade-off exists between the objectives, the fitness is set to $fitness = -\infty$.

\section{Experiments} \label{sec:experiment}

To demonstrate the effectiveness of the proposed method, we conduct experiments to automatically generate benchmark problems that maximize the performance difference between representative pairs of MOEAs.
In this study, we use NSGA-II and MOEA/D as comparison MOEAs for evaluating the generated benchmark problems.
We selected these algorithms because of their different operating principles; NSGA-II is based on Pareto dominance, while MOEA/D is based on decomposition.
For these two algorithms, we consider two experimental scenarios.
In the NSGA-II-preferred scenario, Algorithm A is set to NSGA-II and Algorithm B to MOEA/D.
Conversely, in the MOEA/D-preferred scenario, Algorithm A is set to MOEA/D and Algorithm B to NSGA-II.

\subsection{Parameters of LLM-EBG}

Table~\ref{tab::benchmark_parameter} shows the parameter settings of LLM-EBG in this experiment.
Llama-3-70B is used as an LLM.
The population size $N_{pop}$ is set to 10, and the number of generations $N_{gen}$ is set to 20.
Tournament selection is adopted for parent selection, with a tournament size $T_{size}$ of 3.
The elitism is applied with the number of elites $N_{elites}=1$ to ensure that the individual with the highest fitness is always carried over to the next generation.
The crossover rate $p_{c}$ was set to 0.5; when crossover is not applied, mutation is performed with a probability of $1-p_{c}$.
The number of trials $N_{trials}$ used for evaluating each generated benchmark is set to 20.
Regarding the basic specifications for the benchmark problems, the number of objectives $M_{obj}=2$, the number of decision variables $D=5$, and the search domain is fixed to $[-5,5]^D$.

\subsection{Parameters of MOEAs}

Table~\ref{tab:eval_parameter} shows the parameter settings of MOEAs used internally to evaluate generated benchmarks.
To ensure fair comparison conditions, both NSGA-II and MOEA/D are executed with an identical population size of 100 and a maximum of 100 generations.
We adopt simulated binary crossover (SBX) with a crossover rate of 0.9 and a distribution index of 15, and polynomial mutation (PM) with a mutation rate of $1/D$ (i.e., 0.2) and a distribution index of 20.
As a parameter specific to MOEA/D, the neighborhood size is set to 10, and the Tchebycheff function is used as the scalarizing function.

\begin{table}[tb]
\begin{center}
\caption{Parameters of LLM-EBG}
\label{tab::benchmark_parameter}
\small
\begin{tabular}{lr}
\toprule 
Parameter & Value\\
\midrule
LLM & Llama-3-70B\\
Population size & 10\\
Number of generations & 20\\
Number of elites & 1 \\
Tournament size & 3 \\
Crossover rate & 0.5\\
Number of trials & 20\\
\midrule
Number of objectives $M_{obj}$& 2\\
Number of dimensions $D$ & 5\\
Search range & $[-5, 5]^D$\\
\bottomrule
\end{tabular}
\end{center}
\end{table}

\begin{table}[tb]
\begin{center}
\caption{Parameters of MOEAs for benchmark evaluation}
\label{tab:eval_parameter}
\small
\begin{tabular}{lr}
\toprule 
Parameter & Value \\
\midrule
Search algorithms & NSGA-II / MOEA/D \\
Population size & 100\\
Number of generations & 100 \\
Crossover method & SBX \\
Crossover rate & 0.9 \\
Crossover distribution index & 15 \\
Mutation method & PM \\
Mutation rate & $1/D$ (0.2) \\
Mutation distribution index & 20 \\
\midrule
MOEA/D neighborhood size & 10 \\
MOEA/D scalarizing function & Tchebycheff function \\
\bottomrule
\end{tabular}
\end{center}
\end{table}

\section{Results} \label{sec:result}
This section presents the results of two experiments: the search for NSGA-II-preferred problems and the search for MOEA/D-preferred problems.

We conducted five independent trials for each experiment.
When generating NSGA-II-preferred problems, the maximum fitness value ($fitness=400$) was reached in all five trials.
In contrast, in the experiments for generating MOEA/D-preferred problems, only two out of five trials achieved high fitness ($fitness=400$ or $fitness=226$), while the remaining three trials resulted in $fitness < 0$.
These results indicate that the MOEA/D-preferred scenario is more challenging in the proposed settings of LLM-EBG.
The discrepancy in these success rates is discussed in Section \ref{sec:discussion}.

In the following sections, representative results from the successful trials are reported.

\subsection{NSGA-II-preferred Scenario}
\begin{figure}[t]
\centering
\begin{tcolorbox}[
    colback=white,
    colframe=black!60,
    arc=2mm,
    boxrule=0.8pt,
    boxsep=3pt,
    left=4pt, right=4pt, top=4pt, bottom=4pt,
    fontupper=\scriptsize\ttfamily
  ]
  $f_1(x) = x_0^2 x_2 - 1.92 x_1 x_3 + \sin^2(0.85 x_4) \sqrt{x_0^2 + 1.625 x_2^2}$ \\
  $f_2(x) = \frac{1.26 x_0^2 + 1.035 x_1^2}{1 + \sqrt{0.72 x_2^2 + 1.46 x_3^2}} + 2.6 |0.985 x_4 - 0.63|^2 - 0.99 x_0 x_3$
\end{tcolorbox}
\caption{NSGA-II-preferred problem ($fitness=400$)}
\label{fig:nsga2-preferred-problem}
\centering
\begin{minipage}[t]{0.49\columnwidth}
\centering
\includegraphics[width=\linewidth]{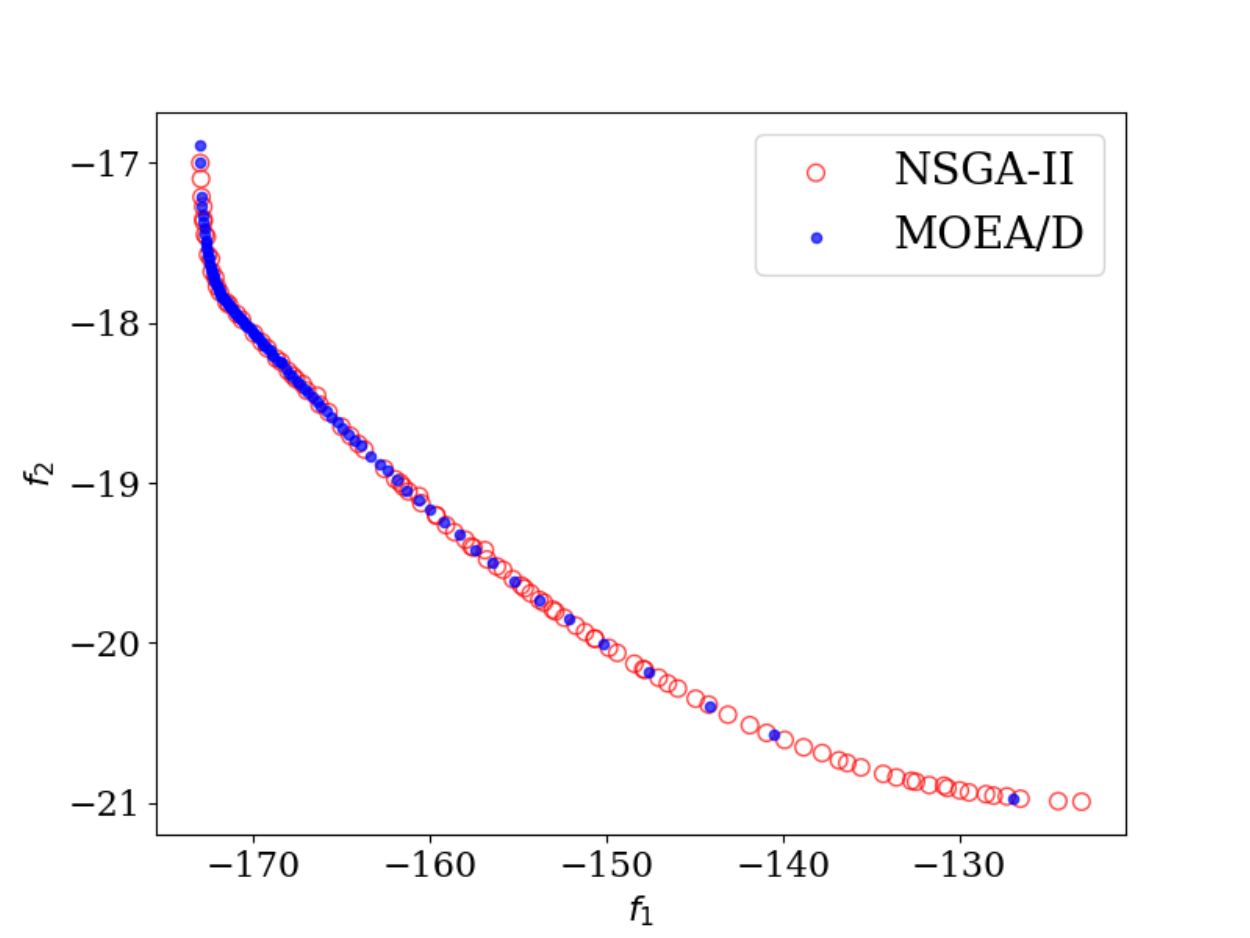}
\subcaption{Pareto front}
\label{fig:result_nsga2_pf}
\end{minipage}
\hfill
\begin{minipage}[t]{0.49\columnwidth}
\centering
\includegraphics[width=\linewidth]{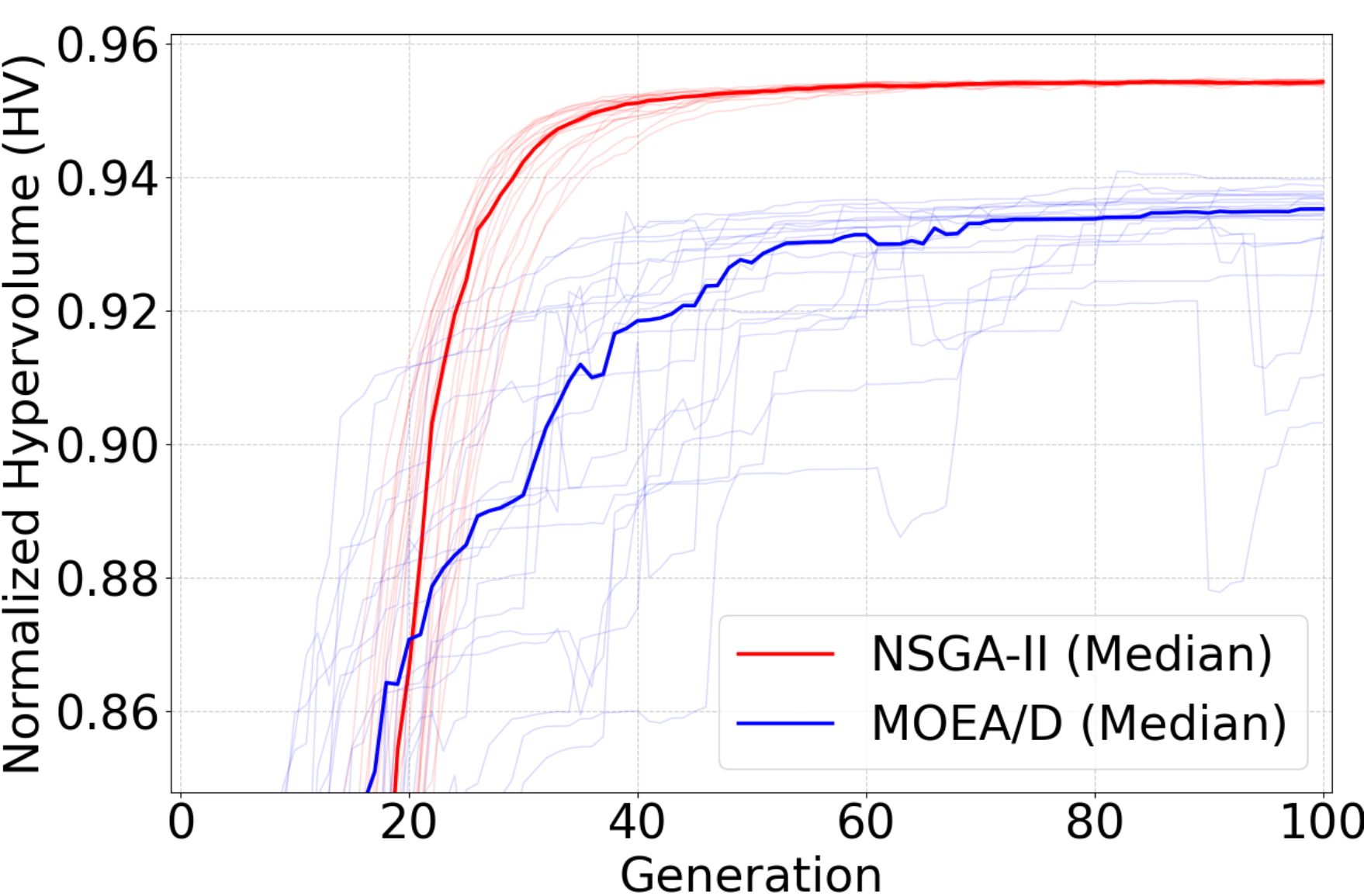}
\subcaption{Transition of HV}
\label{fig:result_nsga2_conv}
\end{minipage}
\caption{Results on NSGA-II-preferred scenario}
\label{fig:result_nsga2}
\end{figure}

Fig.~\ref{fig:nsga2-preferred-problem} shows the benchmark generated in the NSGA-II-preferred scenario in a representative trial, and Fig.~\ref{fig:result_nsga2} shows the results optimizing this benchmark using NSGA-II and MOEA/D.

Fig.~\ref{fig:result_nsga2_pf} shows the Pareto fronts obtained by NSGA-II and MOEA/D.
To illustrate the typical behavior of the algorithms, we plot the non-dominated solutions from the trial that achieved the median Hypervolume (HV) value at the final generation among the 20 independent runs.
This specific trial corresponds to the median trajectory shown in the transition of HV plot (Fig.~\ref{fig:result_nsga2_conv}).
The horizontal and vertical axes denote the first and second objective functions, respectively.
Red open circles indicate the non-dominated solution obtained by NSGA-II, while blue filled circles represent those obtained by MOEA/D.
As shown in Fig.~\ref{fig:result_nsga2_pf}, the solutions obtained by MOEA/D are biased toward the upper left region corresponding to the minimization of the first objective ($f_1$), whereas NSGA-II discovers solutions over a wide range of the Pareto front. 

Fig.~\ref{fig:result_nsga2_conv} shows the HV trajectories over generations for 20 independent trials on this benchmark.
The horizontal axis represents the number of generations, the vertical axis represents HV.
Red and blue lines correspond to NSGA-II and MOEA/D, respectively, and the trials achieving the median HV in the final generation are highlighted with bold lines. 

From Fig.~\ref{fig:result_nsga2_conv}, it is evident that NSGA-II consistently converges to high HV values within the first half of the search (around the 40th generation) with relatively small variations across trials.
In contrast, MOEA/D shows slower convergence and larger variations between trials.
The median HV of NSGA-II (0.9542) clearly exceeds that of MOEA/D (0.9352).

This performance difference is due to the different scales of the objective functions $f_1$ and $f_2$.
As shown in Fig.~\ref{fig:result_nsga2_pf}, the range of $f_1$ spans approximately from $-120$ to $-175$, while $f_2$ ranges from approximately $-17$ to $-21$, resulting in a scale difference of roughly one order of magnitude. 

In this study, MOEA/D employs the Tchebycheff scalarizing function, which minimizes the maximum weighted differences from the ideal point.
When one objective exhibits a much larger scale than the others, the scalarizing function is dominated by that objective, causing the search direction biased toward the single-objective optimization.
Consequently, MOEA/D fails to discover diverse solutions across the entire Pareto front and instead concentrates on minimizing $f_1$ as observed in Fig.~\ref{fig:result_nsga2_pf}. 

In contrast, NSGA-II relies on the Pareto dominance relations, which are invariant to the absolute scale of objective values.
Consequently, NSGA-II remains robust even when objective functions have significantly different scales.
These findings indicate that the proposed framework successfully generated benchmarks with uneven objective scales, thereby exploiting a well-known weakness of MOEA/D.

\subsection{MOEA/D-preferred Scenario}

\begin{figure}[tb]
\centering
\begin{tcolorbox}[
    colback=white,
    colframe=black!60,
    arc=2mm,
    boxrule=0.8pt,
    boxsep=3pt,
    left=4pt, right=4pt, top=4pt, bottom=4pt,
    fontupper=\scriptsize\ttfamily
  ]
\scriptsize{
  $f_1(x) = x_0^2 + 1.4 x_1^2 + 1.01 x_2^2 - 15.1 \sin(0.75 x_3) - 8 |x_4 - 1.6|$ \\
  $f_2(x) = (x_0 + 0.7)^2 + (x_1 + 0.52)^2 + (x_2 + 1.12)^2 + 1.4 x_3^2 + 3.2 x_4^2$ 
  }
\end{tcolorbox}
\caption{MOEA/D-preferred problem ($fitness=400$)}
\label{fig:moead-preferred-problem}
\centering
\begin{minipage}[t]{0.48\columnwidth}
\includegraphics[width=\linewidth]{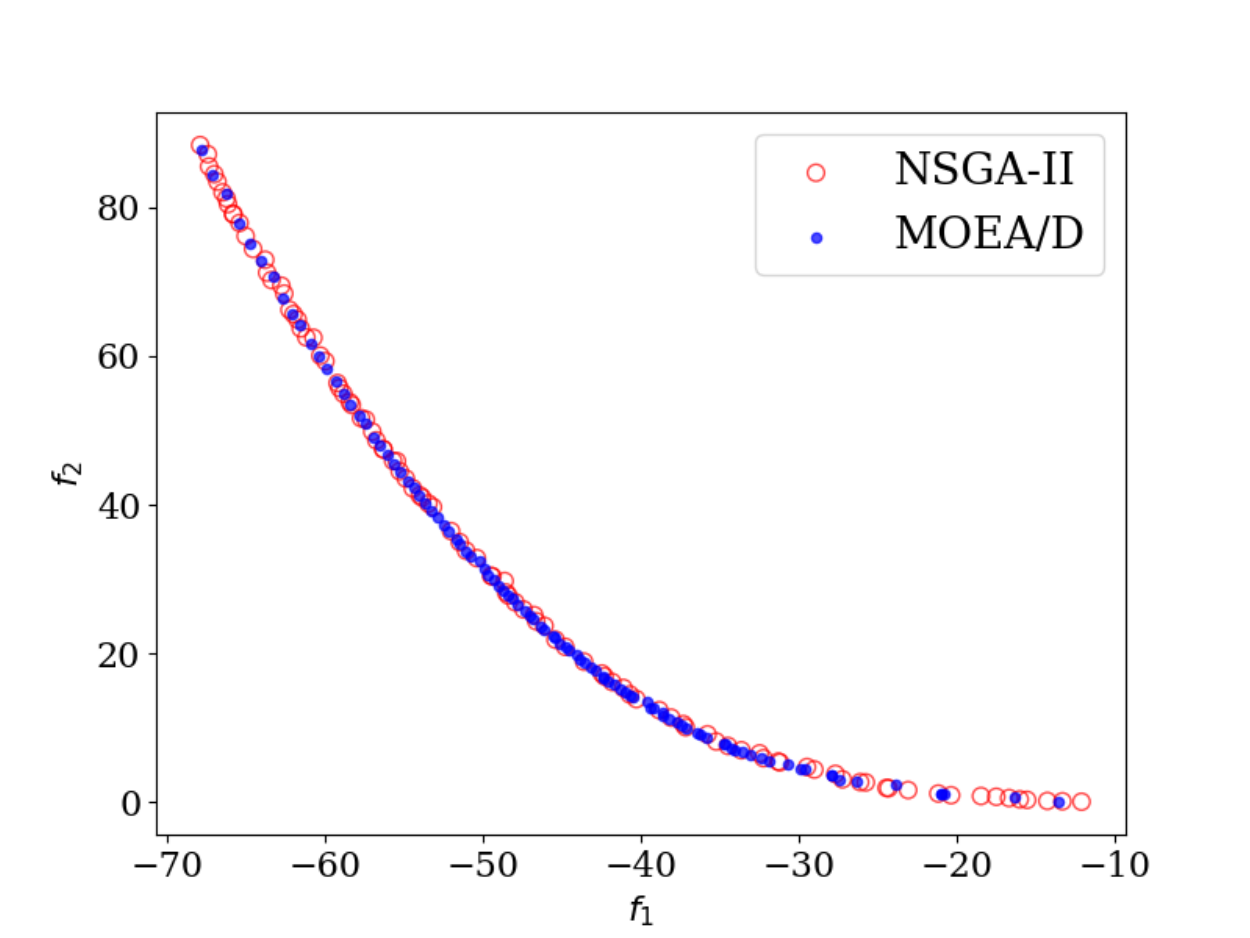}
\subcaption{Pareto front\label{fig:result_moead_pf}}
\end{minipage}
\begin{minipage}[t]{0.48\columnwidth}
\includegraphics[width=\linewidth]{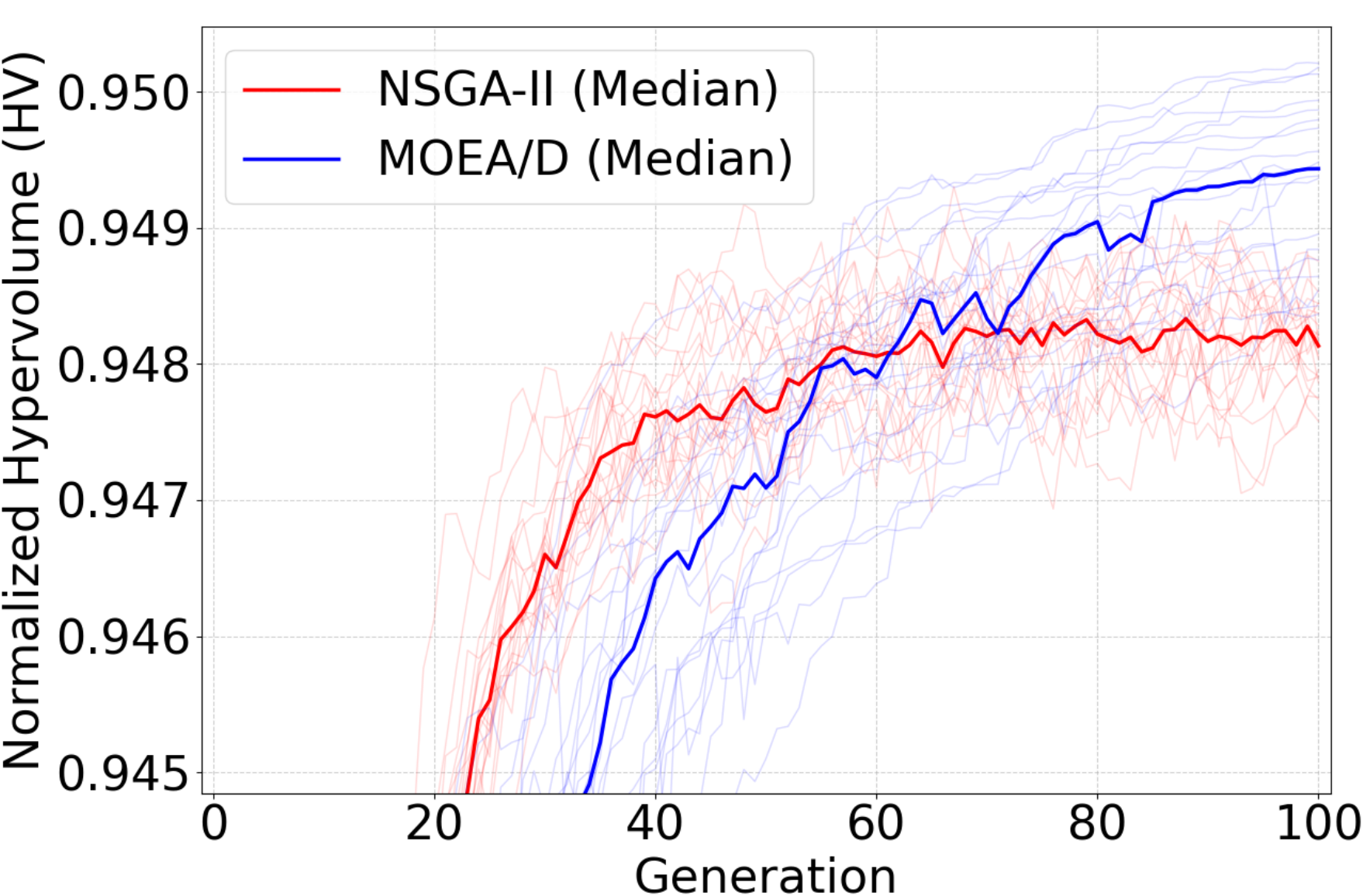}
\subcaption{Transition of HV\label{fig:result_moead_conv}}
\end{minipage}
\caption{Results on MOEA/D-preferred scenario}
\label{fig:result_moead}
\end{figure}
Fig.~\ref{fig:moead-preferred-problem} shows the benchmark generated in the MOEA/D-preferred scenario in a representative trial, and Fig.~\ref{fig:result_moead} illustrates the results optimizing this benchmark using NSGA-II and MOEA/D.

Fig.~\ref{fig:result_moead_pf} compares the Pareto fronts for the MOEA/D-preferred benchmark.
As shown in  Fig.~\ref{fig:result_moead_pf}, while NSGA-II acquires solutions uniformly distributed across the Pareto front, the solutions obtained by MOEA/D are more sparsely distributed near the extremes. 
MOEA/D performs a more precise and dense search in the central region of the Pareto front, which allows it to achieve a higher HV than NSGA-II.

Fig.~\ref{fig:result_moead_conv} shows the HV trajectories for this benchmark.
As in Fig.~\ref{fig:result_nsga2_conv}, red and blue lines correspond to NSGA-II and MOEA/D, respectively, and bold lines indicate the trials achieving the median HV. 

From Fig.~\ref{fig:result_moead_conv}, it can be confirmed that the median HV of MOEA/D (0.9494) is higher than the median HV of NSGA-II (0.9481), although the difference is smaller compared to the NSGA-II-preferred scenario.
NSGA-II converges faster, whereas MOEA/D gradually improves solution quality and ultimately achieves a higher HV.

The Pareto front of this benchmark (Fig.~\ref{fig:result_moead_pf}) has a smooth convex shape.
For such benchmarks, the uniformly distributed reference vectors used by MOEA/D worked advantageously in guiding the search toward a dense and well-distributed approximation of the Pareto front, especially in the central region that contributes significantly to HV improvement. 

In a two-objective problem with convex Pareto fronts, improvements in the central region of the front tend to yield larger gains in HV than improvements near the extremes.
Specifically, a small improvement toward the ideal point in the central region expands the dominated HV substantially, whereas improvements near extreme solutions add only a limited rectangular region to the HV. 

NSGA-II allocates computational resources to preserving extreme solutions that contribute relatively little to HV improvement due to its diversity maintenance mechanism based on crowding distance.
In contrast, MOEA/D focuses search efforts along reference directions, enabling more efficient exploration of the central region of the Pareto front. This characteristic allows MOEA/D to ultimately achieve a higher HV value than NSGA-II in this scenario.

\subsection{Landscape Analysis}
\label{sec:result_feature}

\begin{figure}[tb]
\begin{center}
\includegraphics[width=0.8\linewidth]{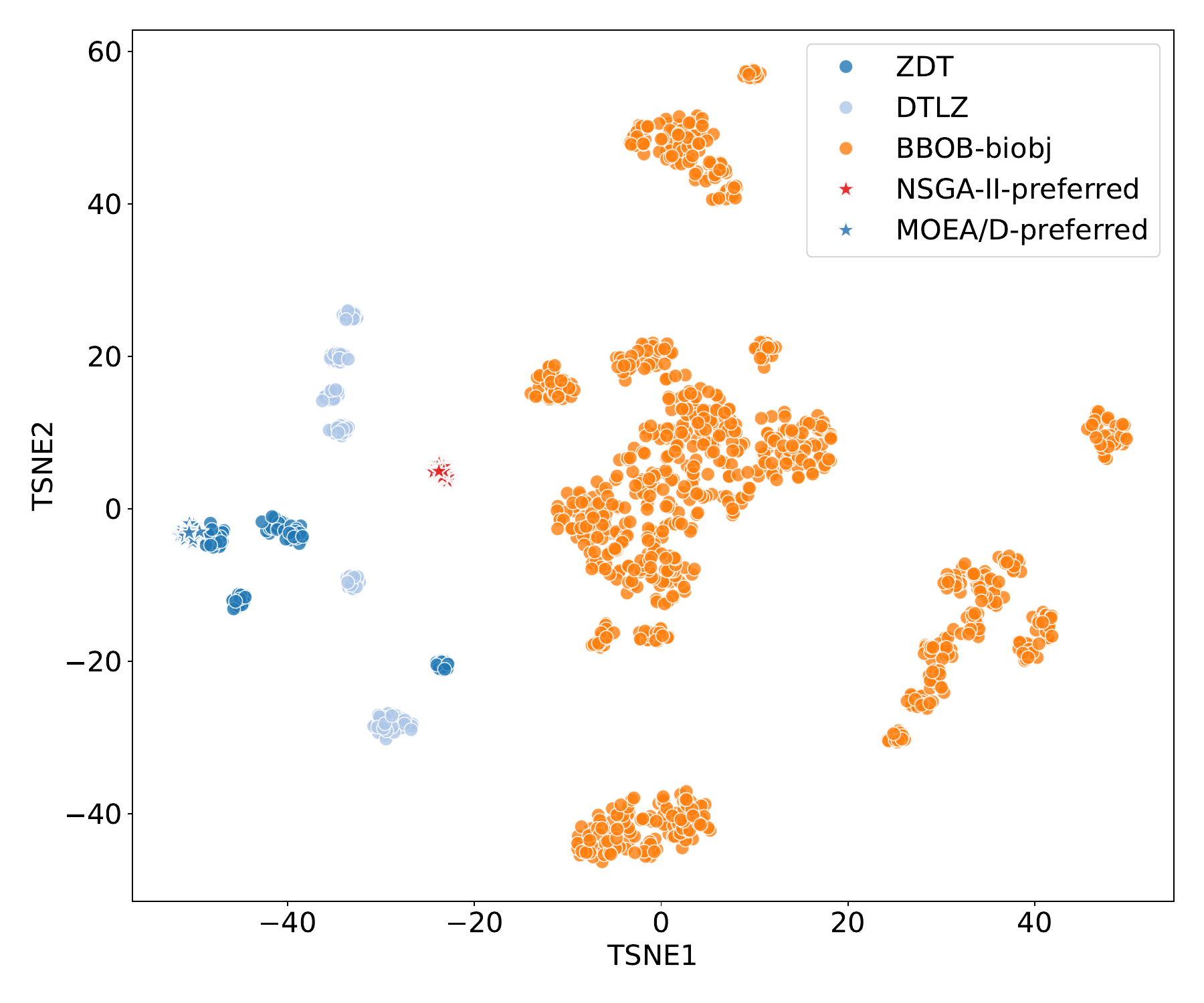}
\caption{Distribution of benchmark problems in the landscape feature space (t-SNE)}
\label{fig:mo_tsne}
\end{center}
\end{figure}
To investigate the landscape characteristics of the generated benchmark problems compared to existing standard benchmark problems, we conducted an analysis using the features proposed by Liefooghe et al. \cite{Liefooghe2021Landscape}.
We use the ZDT (ZDT1--4, 6), DTLZ  (DTLZ1--7), and BBOB-biobj test suites as existing benchmarks.
We calculated the feature vectors for these problems alongside the NSGA-II-preferred and MOEA/D-preferred problems generated by the proposed method.

Fig.~\ref{fig:mo_tsne} shows the results of compressing and visualizing these high-dimensional feature vectors into two dimensions using t-SNE.
As shown in Fig.~\ref{fig:mo_tsne}, the ZDT, DTLZ, and BBOB-biobj benchmark problems form distinct clusters in specific regions of the feature space. Moreover, the MOEA/D-preferred problems are embedded in regions close to the ZDT benchmarks. In contrast, the NSGA-II-preferred problems form clusters that are separated from the existing benchmarks.

These results suggest that existing benchmark suites may underestimate the performance of NSGA-II, as they do not sufficiently cover regions of the feature space where NSGA-II exhibits advantages. This observation highlights the need to incorporate greater diversity into benchmark problem design.

\section{Discussion}
\label{sec:discussion}
As described in Section \ref{sec:result}, the experiments for generating NSGA-II-preferred problems reached the maximum fitness value ($fitness=400$) in all five trials.
In contrast, the experiments to generate MOEA/D-preferred problems achieved high fitness in only 2 out of 5 trials.
We discuss the reasons for this discrepancy below.

This is because NSGA-II-preferred benchmarks were more easily generated under the problem settings in this study.
Specifically, the generation of two-objective problems might have inherently favored NSGA-II, while
MOEA/D tends to show its strength as the number of objectives increases. Therefore, the relative advantage of MOEA/D is comparatively smaller in two-objective problems.
Furthermore, problem structures that exploit MOEA/D’s weaknesses, such as scale differences between objective functions, were easily generated, which influenced the difference in success rates.
In fact, many problems in the initial population exhibited higher HV values under NSGA-II, putting the search for MOEA/D-preferred problems at a disadvantage from the outset.

Another aspect is the prompt design used in this study.
The current prompt is not fully optimized and provides only basic instructions for genetic operations to the LLM. By introducing feedback information, such as past successful cases or fitness trajectories, the LLM can learn effective operational patterns, thereby improving search efficiency.

By addressing these issues, more stable problem generation can be realized in LLM-EBG.

\section{Conclusion}

This study proposed a new framework to automatically generate MOP benchmarks by integrating LLMs with EAs.
In the proposed method, the LLM is employed as a genetic operator for crossover and mutation, evolving benchmark problems represented by combinations of mathematical formulas.
The generated benchmarks are evaluated based on differences in the rank sums of HV across multiple independent optimization runs.
To investigate the effectiveness of the proposed method, we conducted experiments to generate two types of problems that exhibit significant performance differences between NSGA-II and MOEA/D: NSGA-II-preferred and MOEA/D-preferred problems.

The experimental results demonstrated that the proposed framework is capable of evolving problems in which the targeted algorithm outperforms the alternative in terms of search performance.
In particular, NSGA-II-preferred problems exploited MOEA/D's sensitivity to objective scale differences, while MOEA/D-preferred problems featured convex Pareto fronts that favor decomposition-based search. These findings confirm that the proposed method can expose algorithm-specific strengths and weaknesses through automatically generated benchmarks.

As future work, we plan to introduce landscape feature-based evaluation criteria
to explicitly explore benchmarks that are highly novel compared with existing benchmark problems.
In addition, we will extend the proposed framework to many-objective optimization problems (MaOPs) with three or more objectives, as well as to constraint MOPs.

\section*{Acknowledgements}
During the preparation of this work, the authors used AI systems to improve language and readability. After using these tools, the authors reviewed and edited the content as needed and took full responsibility for the publication.

\bibliographystyle{IEEEtran}
\bibliography{reference}

\end{document}